%% file: bias.tex
\pdfoutput=1
\documentclass[11pt]{paper}
\usepackage{geometry}
\usepackage{mathpazo}
\usepackage{amsmath,amsthm,amssymb,xspace}
\usepackage{xspace}
\usepackage{microtype}

\newtheorem{defn}{Definition}[section]

\newtheorem{theorem}{Theorem}[section]
\newtheorem{lemma}{Lemma}[section]
\newtheorem{conjecture}{Conjecture}[section]
\usepackage{url}
\usepackage{mathtools}
\mathtoolsset{showonlyrefs,mathic}
\usepackage{cite}
\usepackage{xcolor}
\usepackage{hyperref}
\usepackage[final,inline,nomargin,index]{fixme}
\fxsetup{theme=colorsig,mode=multiuser,inlineface=\itshape,envface=\itshape}
\FXRegisterAuthor{sv}{asv}{Suresh}
\FXRegisterAuthor{sf}{asf}{Sorelle} 
\FXRegisterAuthor{cs}{acs}{Carlos} 
\FXRegisterAuthor{jm}{ajm}{John} 

\usepackage{graphicx}
\graphicspath{{.}{figures/ggplot/}}

\newcommand{\sens}{\text{sensitivity}}
\newcommand{\specf}{\text{specificity}}
\newcommand{\lrplus}{\ensuremath{\text{LR}_+}}
\newcommand{\di}{\ensuremath{\mathsf{DI}}\xspace}
\newcommand{\ber}{\text{\textsc{ber}}\xspace}

\newcommand{\yes}{\text{YES}}
\newcommand{\no}{\text{NO}}
\newcommand{\innerprod}[2]{\left\langle #1, #2 \right\rangle}

\newcommand{\mypara}[1]{\noindent\textbf{\sffamily #1}}

\title{Certifying and removing disparate impact\thanks{Author ordering on this paper is alphabetical.  This research funded in part by NSF grant 1251049 under the BIGDATA program}}
\author{Michael Feldman\\Haverford College \and Sorelle A. Friedler\\Haverford College \and John Moeller\\University of Utah \and Carlos
  Scheidegger\\University of Arizona \and Suresh
  Venkatasubramanian\thanks{Corresponding author.}\\University of Utah}
\newcommand{\reals}{\mathbb{R}}

\newfont{\mycrnotice}{ptmr8t at 7pt}
\newfont{\myconfname}{ptmri8t at 7pt}

\begin{document}
\maketitle

\input abstract

\input intro

\input prelim

\input certify

\input repair

\input experiments

\input discussion2

\section{Acknowledgments}

This research was funded in part by the  NSF under grant BIGDATA-1251049. Thanks to Deborah Karpatkin, David Robinson, and Natalie Shapero for helping us understand the legal interpretation of disparate impact.  Any misunderstandings about these issues in this paper are our own.  Thanks also to Mark Gould for pointing us to the Griggs v. Duke decision, which helped to set us down this path in the first place.

\bibliographystyle{abbrv}
\bibliography{bias}

\input appendix

\end{document}

%% file: abstract.tex
\begin{abstract}
What does it mean for an algorithm to be biased? 
In U.S. law, unintentional bias is encoded via
\emph{disparate impact}, which occurs when a selection process has widely different outcomes for different groups, even as it appears to be neutral. This legal determination hinges on a definition of a protected class (ethnicity, gender) and an explicit description of the process.

When computers are involved, 
determining disparate impact (and hence bias) is harder. It
might not be possible to disclose the process. In addition,
even if the process is
open, it might be hard to elucidate in a legal setting how the algorithm makes its decisions. Instead of requiring access to the process, we propose making
inferences based on the \emph{data} it uses.

We present four contributions. First, we link disparate impact
to a measure of classification accuracy that while known,
has received relatively little attention.
Second, we propose a test for disparate impact based on
how well the protected class can be predicted from the other attributes.
Third, we describe methods by which data might be made unbiased.  Finally, we present empirical evidence supporting the effectiveness of our test for disparate impact and our approach for both masking bias and preserving \emph{relevant} information in the data.  Interestingly, our approach resembles some actual selection practices that have recently received legal scrutiny.

\end{abstract}


%% file: intro.tex
\section{Introduction}

In Griggs v. Duke Power Co.~\cite{griggs1971}, the US Supreme Court ruled a business hiring decision illegal if it resulted in disparate impact by race even if the decision was not explicitly determined based on race.  The Duke Power Co. was forced to stop using intelligence test scores and high school diplomas, qualifications largely correlated with race, to make hiring decisions. The Griggs decision gave birth to the legal doctrine of \emph{disparate impact}, which today is the predominant legal theory used to determine \emph{unintended} discrimination in the U.S. Note that disparate \emph{impact} is different from disparate \emph{treatment}, which refers to \emph{intended} or direct discrimination.  Ricci v. DeStefano \cite{Ricci09} examined the relationship between the two notions, and disparate impact remains a topic of legal interest.

\looseness-1 Today, algorithms are being used to make decisions both large and small in almost all aspects of our lives, whether they involve mundane tasks like recommendations for buying goods, predictions of credit rating prior to approving a housing loan, or even life-altering decisions like sentencing guidelines after conviction\cite{Hodson15NewScientist}. How do we know if these algorithms are biased, involve illegal discrimination, or are unfair? 

%
These concerns have generated calls, by governments and NGOs alike, for research into these issues~\cite{podesta14bigdata, civilRightsPrinciples}.
In this paper, we introduce and address two such problems with the goals of quantifying and then removing disparate impact.

While the Supreme Court has resisted a ``rigid mathematical formula'' defining disparate impact~\cite{watson1988}, we will adopt a generalization of the 80 percent rule advocated by the US Equal Employment Opportunity Commission (EEOC)~\cite{eeoc1979}. 
We note that disparate impact itself is not illegal; in hiring decisions, business necessity arguments can be made to excuse disparate impact.
\begin{defn}[{\small Disparate Impact (``80\% rule'')}]
Given data set $D = (X, Y, C)$, with \emph{protected attribute} $X$ (e.g., race, sex, religion, etc.), remaining attributes $Y$, and binary class to be predicted $C$ (e.g., ``will hire''), we will say that $D$ has \emph{disparate impact}
if 
\[ \frac{\Pr(C = YES| X = 0)}{\Pr(C = YES| X = 1)} \leq \tau = 0.8 \]
for positive outcome class $YES$ and majority protected attribute $1$ where $\Pr(C = c | X = x)$ denotes the conditional probability (evaluated over $D$) that the class outcome is $c \in C$ given protected attribute $x \in X$.\footnote{Note that under this definition disparate impact is determined based on the given data set and decision outcomes. Notably, it does not use a broader sample universe, and does not take into account statistical significance as has been advocated by some legal scholars~\cite{peresie09law}.}
\end{defn}

\looseness-1 The two problems we consider address identifying and removing disparate impact.  The \emph{disparate impact certification problem} is to guarantee that, given $D$, any classification algorithm aiming to predict some $C'$ (which is potentially different from the given $C$) from $Y$ would not have disparate impact.  By certifying any outcomes $C'$, and not the process by which they were reached, we follow legal precedent in making no judgment on the algorithm itself, and additionally ensure that potentially sensitive algorithms remain proprietary.  The \emph{disparate impact removal problem} is to take some data set $D$ and return a data set $\bar{D} = (X, \bar{Y}, C)$ that can be certified as not having disparate impact.  The goal is to change only the remaining attributes $Y$, leaving $C$ as in the original data set so that the ability to classify can be preserved as much as possible.

\subsection{Results}

We have four main contributions.

We first introduce these problems to the computer science community and develop its theoretical underpinnings.  The study of the EEOC's 80\% rule as a specific class of loss function does not appear to have received much attention in the literature.  We link this measure of disparate impact to the balanced error rate (\ber).
We show that any decision exhibiting disparate impact can be converted into one where the protected attribute leaks, i.e. can be predicted with low \ber.

Second, this theoretical result gives us a procedure for certifying the impossibility of disparate impact on a data set.  This procedure involves a particular regression algorithm which minimizes \ber. We connect \ber to disparate impact in a variety of settings (point and interval estimates, and  distributions). We discuss these two contributions in Sections \ref{sec:note-meas-disp} and \ref{sec:main-idea}.

In Section \ref{sec:repairing-data}, we show how to transform the input dataset so that predictability of the protected attribute is impossible. We show that this transformation still preserves much of the signal in the unprotected attributes and has nice properties in terms of closeness to the original data distribution.

Finally, we present a detailed empirical study in Section \ref{sec:experiments}.  We show that our algorithm certifying lack of disparate impact on a data set is effective, such that with the three classifiers we used certified data sets don't show disparate impact.  We demonstrate the fairness / utility tradeoff for our partial repair procedures.  Comparing to related work, we find that for any desired fairness value we can achieve a higher accuracy than other fairness procedures.  This is likely due to our emphasis on changing the data to achieve fairness, thus allowing any strong classifier to be used for prediction.

Our procedure for detecting disparate impact goes through an actual classification algorithm. As we show in our experiments, a better classifier provides a more sensitive detector.
We believe this is notable. As algorithms get better at learning patterns, they become more able to introduce subtle biases into the decision-making process by finding subtle dependencies among features. But this very sophistication helps \emph{detect} such biases as well via our procedure! Thus, data mining can be used to verify the fairness of such algorithms as well.

\section{Related Work}
There is, of course, a long history of legal work on disparate impact.  There is also related work under the name \emph{statistical discrimination} in Economics.  We will not survey such work here.  Instead, we direct the reader to the survey of Romei and Ruggieri~\cite{Romei13Multidisciplinary} and to a discussion of the issues specific to data mining and disparate impact~\cite{Barocas14DisparateImpact}.  Here, we focus on data mining research relating to  combating discrimination.  This research can be broadly categorized in terms of methods that achieve fairness by modifying the classifiers and those that achieve fairness by modifying data. 

 Kamishima et al.~\cite{Kamishima12Fairness, Kamishima11Fairness} develop a regularizer for classifiers to penalize prejudicial outcomes and show that this can reduce indirect prejudice (their name for implicit discrimination like disparate impact) while still allowing for accurate classification.  They note that as prejudicial outcomes are decreased, the classification accuracy is also decreased.
Our work falls into the category of algorithms that change the input data.  Previous work has focused on changing the class values of the original data in such a way so that the total number of class changes is small~\cite{Kamiran09Classifying, Calders09Building}, while we will keep the class values the same for training purposes and change the data itself.  Calders et al.~\cite{Calders09Building} have also previously examined one method for changing the data in which different data items are given weights and the weights are adjusted to achieve fairness.  In this category of work, as well, there is worry that the change to the data will decrease the classification accuracy, and Calders et al. have formalized this as a fairness/utility tradeoff~\cite{Calders09Building}.  We additionally note that lower classification accuracy may actually be the desired result, if that classification accuracy was due to discriminatory decision making in the past.

An important related work is the approach of ``fairness through awareness'' of Dwork et al.~\cite{Dwork12Fairness} and Zemel et al.~\cite{icml2013_zemel13}. Dwork et al.~\cite{Dwork12Fairness} focus on the problem of \emph{individual} fairness; their approach posits the existence of a similarity measure between individual entities and seeks to find classifiers that ensure similar outcomes on individuals that are similar, via a Lipschitz condition. In the work of Zemel et al.~\cite{icml2013_zemel13}, this idea of protecting individual fairness is combined with a statistical group-based fairness criterion that is similar to the approach we take in this work. A key contribution of their work is that they learn a modified representation of the data in which fairness is ensured while attempting to preserve fidelity with the original classification task.  While this group fairness measure  is similar to ours in spirit, it does not match the legal definition we base our work on. Another paper that also (implicitly) defines fairness on an individual basis is the work by Thanh et al.~\cite{luong2011knn}. Their proposed repair mechanism changes class attributes of the data (rather than the data itself). 

Pedreschi, Ruggieri and Turini \cite{pedreschi2012study,pedreschi2009integrating} have examined the ``80\% rule'' that we study in this paper as part of a larger class of measures based on a classifier's confusion matrix.

%% file: prelim.tex
\section{Disparate Impact and Error Rates}
\label{sec:note-meas-disp}

We start by reinterpreting the  ``80\% rule'' in terms of more standard statistical measures of quality of a classifier. This presents notational challenges. The terminology of ``right'' and ``wrong'', ``positive'' and ``negative'' that is used in classification is an awkward fit when dealing with majority and minority classes, and selection decisions. For \emph{notational convenience only}, we will use the convention that the protected class $X$ takes on two values: $X =0$ for the ``minority'' class and $X = 1$ for the ``default'' class. For example, in most gender-discrimination scenarios the value $0$ would be assigned to ``female'' and $1$ to ``male''. We will denote a successful binary classification outcome $C$ (say, a hiring decision) by $C =\,$\textsc{yes} and a failure by $C=\, $\textsc{no}. Finally, we will map the majority class to ``positive'' examples and the minority class to ``negative'' examples with respect to the classification outcome, all the while reminding the reader that this is merely a convenience to do the mapping, and does not reflect any judgments about the classes. The advantage of this mapping is that it renders our results more intuitive: a classifier with high ``error'' will also be one that is least biased, because it is unable to distinguish the two classes.

Table \ref{tab:confusion} describes the \emph{confusion matrix} for a classification with respect to the above attributes where each entry is the probability of that particular pair of outcomes for data sampled from the input distribution (we use the empirical distribution when referring to a specific data set).
\begin{table}[htbp]
  \centering
  \begin{tabular}{c|c|c|}
  Outcome & $X = 0$ & $X = 1$ \\ \hline
  $C=\,$\textsc{no} & $a$ & $b$\\ \hline
$C=\,$\textsc{yes} & $c$& $d$\\ \hline
\end{tabular}
\caption{A confusion matrix\label{tab:confusion}}
\end{table}

The $80\%$ rule can then be quantified as: 
\[ \frac{c/(a + c)}{d/(b + d)} \ge 0.8 \]
Note that the traditional notion of ``accuracy'' includes terms in the numerator from both columns, and so cannot be directly compared to the $80\%$ rule. Still, other class-sensitive error metrics are known, and more directly relate to the $80\%$ rule:

\begin{defn}[{\small Class-conditioned error metrics}]
  The \emph{sensitivity} of a test (informally, its true positive rate) is defined as the conditional probability of returning \textsc{yes} on ``positive'' examples (a.k.a. the majority class). In other words, 
\[ \sens = \frac{d}{b+d} \]
  The \emph{specificity} of a test (its true negative rate) is defined as the conditional probability of returning \textsc{no} on ``negative'' examples (a.k.a. the minority) class. I.e.,
\[ \specf = \frac{a}{a+c} \]
\end{defn}

\begin{defn}[{\small Likelihood ratio (positive)}]
The likelihood ratio positive, denoted by \lrplus, is given by 
\[ \lrplus(C, X) = \frac{\sens}{1 - \specf} = \frac{d/(b+d)}{c/(a+c)} \]
\end{defn}

We can now restate the $80\%$ rule in terms of a data set.
\begin{defn}[Disparate Impact]
A data set has \emph{disparate impact} if
\[ \lrplus(C, X) > \frac{1}{\tau} = 1.25 \]
\end{defn}

It will be convenient to work with the reciprocal of $\lrplus$, which we denote by 
\[ \di = \frac{1}{\lrplus(C, X)} \mbox{ .}\]
This will allow us to discuss the value associated with disparate impact before the threshold is applied.



\looseness-1\vspace*{0.1in}\mypara{Multiple classes.} Disparate impact is defined only for two classes. In general, one might imagine a multivalued class attribute (for example, like ethnicity). In this paper, we will assume that a multivalued class attribute has one value designated as the ``default'' or majority class, and will compare each of the other values pairwise to this default class. While this ignores zero-sum effects between the different class values, it reflects the current binary nature of legal thought on discrimination. A more general treatment of joint discrimination among multiple classes is beyond the scope of this work.


%% file: certify.tex
\section{Computational Fairness}
\label{sec:main-idea}

Our notion of computational fairness starts with two players, Alice and Bob. Alice runs an algorithm $\mathcal{A}$ that makes decisions based on some input.  For example, Alice may be an employer using $\mathcal{A}$ to decide who to hire. Specifically, $\mathcal{A}$ takes a data set $D$ with protected attribute $X$ and unprotected attributes $Y$ and makes a (binary) decision $C$. By law, Alice is not allowed to use $X$ in making decisions, and claims to use only $Y$. It is Bob's job to verify that on the data $D$, Alice's algorithm $\mathcal{A}$ is not liable for a claim of disparate impact. 

\vspace*{0.1in}\mypara{Trust model.} We assume that Bob does not have access to
$\mathcal{A}$. Further, we assume that Alice has good intentions: specifically,
that Alice is not secretly using $X$ in $\mathcal{A}$ while lying about
it. While assuming Alice is lying about the use of $X$ might be more plausible, it is much harder to detect. More importantly, from a functional perspective, it does not matter whether Alice uses $X$ explicitly or uses proxy attributes $Y$ that have the same effect: this is the core message from the Griggs case that introduced the doctrine of disparate impact. In other words, our \emph{certification} process is indifferent to Alice's intentions, but our \emph{repair} process will assume good faith.

We summarize our main idea with the following intuition:
\begin{quote}
\emph{  If Bob cannot predict $X$ given the other attributes of $D$, then $\mathcal{A}$ is fair with respect to Bob on $D$. 
}\end{quote}


\subsection{Predictability and Disparate Impact}

We now present a formal definition of predictability and link it to the legal notion of disparate impact.  Recall that $D = (X,Y,C)$ where $X$ is the protected attribute, $Y$ is the remaining attributes, and $C$ is the class outcome to be predicted.

The basis for our formulation is a procedure that predicts $X$ from $Y$. We would like a way to measure the quality of this predictor in a way that a) can be optimized using standard predictors in machine learning and b) can be related to $\lrplus$. The standard notions of accuracy of a classifier fail to do the second (as discussed earlier) and using $\lrplus$ directly fails to satisfy the first constraint. 

The error measure we seek turns out to be the \emph{balanced error rate} \ber. 

\begin{defn}[BER]
  Let $f : Y \to X$ be a predictor of $X$ from $Y$. The \emph{balanced error rate} \ber of $f$ on distribution $\mathcal{D}$ over the pair $(X, Y)$ is defined as the (unweighted) average class-conditioned error of $f$. In other words, 
\[ \ber(f(Y), X) = \frac{\Pr[f(Y) = 0 | X = 1] + \Pr[f(Y) = 1 | X = 0] }{2} \]
\end{defn}

\begin{defn}[{\small Predictability}]
$X$ is said to be \emph{$\epsilon$-predictable} from $Y$ if there exists a function $f:Y \rightarrow X$ such that
\[\ber(f(Y), X) \le\epsilon. \]
\end{defn}

This motivates our definition of $\epsilon$-fairness, as a data set that is \emph{not} predictable.
\begin{defn}[$\epsilon$-fairness]
A data set $D = (X, Y, C)$ is said to be \emph{$\epsilon$-fair} if for any
classification algorithm $f \colon Y \to X$ 
\[\ber(f(Y), X) > \epsilon \]
with (empirical) probabilities estimated from $D$.
\end{defn}

Recall the definition of disparate impact from Section \ref{sec:note-meas-disp}.  We will be interested in examining the potential disparate impact of a classifier $g \colon Y \rightarrow C$ and will consider the value $\di(g) = 1 / \lrplus(g(Y), X)$ as it relates to the threshold $\tau$.  Where $g$ is clear from context, we will refer to this as $\di$.

The justification of our definition of fairness comes from the following theorem:
\begin{theorem}
\label{sec:conn-pred-disp-1}
A data set is $(1/2-\beta/8)$-predictable if and only if it admits disparate impact, where $\beta$ is the fraction of elements in the minority class $(X = 0)$ that are selected $(C = 1)$. 
\end{theorem}

\begin{proof}
We will start with the direction showing that disparate impact implies predictability.  Suppose that there exists some function $g:Y \rightarrow C$ such that $\lrplus(g(y), c) \geq \frac{1}{\tau}$.  We will create a function $\psi:C \rightarrow X$ such that $\ber(\psi(g(y)), x) < \epsilon$ for $(x,y) \in D$. Thus the combined predictor $f = \psi \circ g$ satisfies the definition of predictability. 

Consider the confusion matrix associated with $g$, depicted in \autoref{tab:di-confusion}.
\begin{table}[htbp]
  \centering
  \begin{tabular}{r|c|c}
   Prediction & $X = 0$ & $X = 1$ \\ \hline
   $g(y) = \no  $ & a       & b       \\
   $g(y) = \yes $ & c       & d
  \end{tabular}
  \caption{Confusion matrix for $g$\label{tab:di-confusion}}
\end{table}
Set $\alpha \triangleq
\frac{b}{b+d}$ and $\beta\triangleq \frac{c}{a+c}$. Then we can write 
$ \lrplus(g(y), X) = \frac{1-\alpha}{\beta}$ and $\di(g) = \frac{\beta}{1 - \alpha}$. 

We define the \emph{purely biased} mapping $\psi \colon C \to X$ as $\psi(\yes) = 1$ and $\psi(\no) = 0$. Finally, let $\phi \colon Y \to X = \psi \circ g$. The confusion matrix for $\phi$ is depicted in \autoref{tab:ber-confusion}. Note that the confusion matrix for $\phi$ is identical to the matrix for $g$. 
\begin{table}[htbp]
  \centering
  \begin{tabular}{r|c|c}
   Prediction & $X = 0$ & $X = 1$ \\ \hline
   $\phi(Y) = 0  $ & a       & b       \\
   $\phi(Y) = 1  $ & c       & d
  \end{tabular}
  \caption{Confusion matrix for $\phi$\label{tab:ber-confusion}}
\end{table}

We can now express $\ber(\phi)$ in terms of this matrix. Specifically, 
$ \ber(\phi) = \frac{\alpha + \beta}{2}$.

\vspace*{0.1in}
\mypara{Representations.} 
We can now express contours of the $\di$ and $\ber$ functions as curves in the unit square $[0,1]^2$. 
Reparametrizing $\pi_1 = 1 - \alpha$ and $\pi_0 = \beta$, we can express the error
measures as $  \di(g) =  \frac{\pi_0}{\pi_1} $ and $ \ber(\phi) = \frac{1 + \pi_0 - \pi_1}{2} $

As a consequence, any classifier $g$ with $\di(g) = \delta$ can be represented
in the $[0,1]^2$ unit square as the line $\pi_1 = \pi_0/\delta$. Any classifier $\phi$
with $\ber(\phi) = \epsilon$ can be written as the function $\pi_1 = \pi_0 + 1 - 2\epsilon$.

Let us now fix the desired \di\ threshold $\tau$, corresponding to the line $\pi_1
=\pi_0/\tau$. Notice that the region $\{(\pi_0,\pi_1)\mid \pi_1 \ge \pi_0/\tau\}$ is the region
where one would make a finding of disparate impact (for $\tau = 0.8$).

Now given a classification that admits a finding of disparate impact, we can
compute $\beta$. Consider the point $(\beta, \beta/\tau)$ at which the line $\pi_0 = \beta$ intersects the
\di curve $\pi_1 = \pi_0/\tau$. This point lies on the \ber contour $(1 +
\beta-\beta/\tau)/2 = \epsilon$, yielding 
$ \epsilon = 1/2 - \beta(\frac{1}{\tau}-1)/2 $
In particular, for the \di threshold of $\tau = 0.8$, the desired \ber threshold is
\[\epsilon = \frac{1}{2} - \frac{\beta}{8} \]
and so disparate impact implies predictability.

With this infrastructure in place, the other direction of the proof is now easy.  To show that predictability implies disparate impact, we will use the same idea of a purely biased classifier.  Suppose there is a function $f \colon Y \to X$ such that $\ber(f(y), x) \leq \epsilon$.  Let $\psi^{-1} \colon X \to C$ be the inverse purely biased mapping, i.e. $\psi^{-1}(1) = \yes$ and $\psi^{-1}(0) = \no$.  Let $g \colon Y \to C = \psi^{-1} \circ f$.  Using the same representation as before, this gives us
$ \pi_1 \geq 1 + \pi_0 - 2  \epsilon$ and therefore
\[ \frac{\pi_0}{\pi_1} \leq \frac{\pi_0}{1 + \pi_0 - 2  \epsilon} = 1 - \frac{1 - 2 \epsilon}{\pi_0 + 1 - 2 \epsilon} \]
Recalling that $\di(g) = \frac{\pi_0}{\pi_1}$ and that $\pi_0 = \beta$ yields
$ \di(g) \leq 1 - \frac{1 - 2 \epsilon}{\beta + 1 - 2\epsilon} = \tau$.
For $\tau = 0.8$, this again gives us a desired \ber threshold of
$\epsilon = \frac{1}{2} - \frac{\beta}{8}$.
\end{proof}

Note that as $\epsilon$ approaches $1/2$ the bound tends towards
the trivial (since any binary classifier has \ber at most $1/2$). In
other words, as $\beta$ tends to $0$, the bound becomes vacuous. 

This points to an interesting line of attack to evade a disparate impact finding. Note that
$\beta$ is the (class conditioned) rate at which members of the protected class
are selected. Consider now a scenario where a company is being investigated for
discriminatory hiring practices. One way in which the company might defeat such
a finding is by interviewing (but not hiring) a large proportion of applicants
from the protected class. This effectively drives $\beta$ down, and the
observation above says that in this setting their discriminatory practices will
be harder to detect, because our result can not guarantee that a classifier will
have error significantly less than $0.5$. 

Observe that in this analysis we use an extremely weak classifier to prove the
existence of a relation between predictability and disparate impact. It is
likely that using a better classifier (for example the Bayes optimal classifier or even a classifier that optimizes \ber) might yield a stronger relationship between the two notions. 

\vspace*{0.1in}\mypara{Dealing with uncertainty.} In general, $\beta$ might be
hard to estimate from a fixed data set, and in practice we might only know that
the true value of $\beta$ lies in a range $[\beta_\ell, \beta_u]$. Since the
\ber threshold varies monotonically with $\beta$, we can merely use $\beta_\ell$
to obtain a conservative estimate. 

Another source of uncertainty is in the \ber estimate itself. Suppose that our
classifier yields an error that lies in a range $[\gamma, \gamma']$. Again,
because of monotonicity, we will obtain an interval of values $[\tau, \tau']$
for \di. Note that if (using a Bayesian approach) we are able to build a
distribution over \ber, this distribution will then transfer over to the \di
estimate as well. 

\subsection{Certifying (lack of) DI with SVMs}
\label{sec:certifying-lack-of}

The above argument gives us a way to determine whether a data set is potentially
amenable to disparate impact (in other words, whether there is insufficient
information to detect a protected attribute from the provided data).

\looseness-1 \vspace*{0.1in}\mypara{Algorithm.}
We run a classifier that optimizes \ber on the given data set, attempting to
predict the protected attributes $X$ from the remaining attributes $Y$.  Suppose
the error in this prediction is $\epsilon$. Then using the estimate of $\beta$
from the data, we can substitute this into the equation above and obtain a
threshold $\epsilon'$.  If $\epsilon' < \epsilon$, then we can declare the data
set free from disparate impact. 

\looseness-1 Assume that we have an optimal classifier with respect to \ber. Then we know
that all classifiers will incur a \ber of at least $\epsilon$. By Theorem
\ref{sec:conn-pred-disp-1}, this implies that no classifier on $D$ will exhibit
disparate impact, and so our certification is correct. 

The only remaining question is what classifier is used by this algorithm.  The usual way to incorporate class sensitivity into a classifier is to use different costs for misclassifying points in different classes. A number of class-sensitive cost measures fall into this framework, and there are algorithms for optimizing these measures (see \cite{menon2013statistical} for a review), as well as a general (but expensive) method due to Joachims that does a clever grid search over a standard SVM to optimize a large family of class-sensitive measures\cite{joachims2005support}. Oddly, \ber is not usually included among the measures studied. 

Formally, as pointed out by Zhao et al\cite{zhao2013beyond}, \ber is not a
cost-sensitive classification error measure because the weights assigned to
class-specific misclassification depend on the relative class sizes (so they can
be normalized). However, for any given data set we know the class sizes and can
reweight accordingly. We adapt a standard hinge-loss SVM to incorporate
class-sensitivity and optimize for (regularized) \ber. This adaptation is
standard, and yields a cost function that can be optimized using AdaBoost. 

  We illustrate this by showing how to adapt a standard hinge-loss SVM to instead optimize \ber\footnote{This approach in general is well known; we provide a detailed derivation here for clarity.}. Consider the standard soft-margin (linear) SVM:
 \begin{equation}
 \min_{\vec{w},\boldsymbol{\xi},b}\frac{1}{2}\|\vec{w}\|^2 + \frac{C}{n}\sum_{j=1}^n\xi_j
 \end{equation}
such that
 \begin{equation} 
 y_j(\innerprod{\vec{w}}{\vec{x}_j} + b)\geq 1-\xi_j\quad\text{and}\quad \xi_j\geq 0\label{eq:1}.
 \end{equation}
 The constraints in \eqref{eq:1} are equivalent to the following:
 \begin{equation}
 \xi_j \geq \max\{1 - y_j(\innerprod{\vec{w}}{\vec{x}_j} + b), 0\},
 \end{equation}
 in which the right side is the hinge loss.

 Minimizing the \emph{balanced error rate} (BER) would result in the following optimization:
 \begin{equation}
   \min_{\vec{w},b}\frac{1}{2}\|\vec{w}\|^2 + 
   C\left( 
     \frac{1}{2n^+}\sum_{\{j|y_j=1\}}(1-\hat{y}_j) + 
     \frac{1}{2n^-}\sum_{\{j|y_j=-1\}}(1+\hat{y}_j) 
   \right),
 \end{equation}
 where $n^+ = |\{j\mid y_j=1\}|$ and $n^- = |\{j\mid y_j=-1\}|$ (we'll use $n_j$ to refer to the correct count for the corresponding $y_j$). 
 This optimization is NP-hard for the same reasons as minimizing the 0-1 loss is, so we will relax it to hinge loss as usual:
 \begin{align}
   &\min_{\vec{w},b}\ \frac{1}{2}\|\vec{w}\|^2 + 
   C\left( 
     \frac{1}{2n^+}\sum_{\{j|y_j=1\}}\xi_j + 
     \frac{1}{2n^-}\sum_{\{j|y_j=-1\}}\xi_j 
   \right),
 \\=&\min_{\vec{w},b}\ \frac{1}{2}\|\vec{w}\|^2 + 
   \frac{C}{n} \sum_j\frac{n}{2n_j}\xi_j
 \\=&\min_{\vec{w},b}\ \frac{1}{2}\|\vec{w}\|^2 + 
   \frac{C}{n} \sum_j D_j\xi_j,\label{eq:2}
 \end{align}
 where $\xi_j$ has the same constraints as in \eqref{eq:1}. 
 \eqref{eq:2} now has the same form as an SVM in AdaBoost, so we can use existing techniqes to solve this SVM. 
 Note also that $\sum_j D_j = 1$, making the $D_j$ a distribution over $[1..n]$ -- this means that we can use an AdaBoost formulation without needing to adjust constants.




%% file: repair.tex
\section{Removing disparate impact}
\label{sec:repairing-data}

Once Bob's certification procedure has made a determination of (potential) disparate impact on $D$, Alice might request a \emph{repaired} version $\bar{D}$ of $D$, where any attributes in $D$ that could be used to predict $X$ have been changed so that $\bar{D}$ would be certified as $\epsilon$-fair.  
We now describe how to construct such a set  $\bar{D} = (X, \bar{Y}, C)$ such that $\bar{D}$ does not have disparate impact in terms of protected attribute $X$.  While for notational simplicity we will assume that $X$ is used directly in what follows, in practice the attribute used to stratify the data for repair need not directly be the protected attribute or even a single protected attribute.  In the case of the Texas Top 10\% Rule that admits the top ten percent of every high school class in Texas to the University of Texas \cite{Texas97Top10Percent}, the attribute used to stratify is the high school attended, which is an attribute that correlates with race.  
If repair of multiple protected attributes is desired, the joint distribution can be used to stratify the data.  (We will look into the effects of this experimentally in Section \ref{sec:fairness_utility}.)

Of course, it is important to change the data in such a way that predicting the class is still possible.  Specifically, our goal will be to preserve the relative per-attribute ordering as follows.  Given protected attribute $X$ and a \emph{single} numerical attribute $Y$, let $Y_x = \Pr(Y|X = x)$ denote the marginal distribution on $Y$ conditioned on $X = x$.  Let $F_x: Y_x \to [0,1]$ be the cumulative distribution function for values $y \in Y_x$ and let $F^{-1}_x: [0,1] \to Y_x$ be the associated quantile function (i.e $F^{-1}_x(1/2)$ is the value of $y$ such that $\Pr(Y \ge y|X = x) = 1/2$).
We will say that $F_x$ \emph{ranks} the values of $Y_x$.

Let $\bar{Y}$ be the repaired version of $Y$ in $\bar{D}$.  We will say that $\bar{D}$ \emph{strongly preserves rank} if for any $y \in Y_x$ and $x \in X$, its ``repaired'' counterpart $\bar{y} \in \bar{Y}_x$ has $F_x(y) = F_x(\bar{y})$. Strongly preserving rank in this way, despite changing the true values of $Y$, appears to allow Alice's algorithm to continue choosing stronger (higher ranked) applicants over weaker ones. We present experimental evidence for this in Section~\ref{sec:experiments}.

With this motivation, we now give a repair algorithm that strongly preserves rank and ensures that $\bar{D} = (X, \bar{Y}, C)$ is fair (i.e., is $\epsilon$-fair for $\epsilon = 1/2$).  In the discussion that follows, for the sake of clarity we will treat $Y$ as a single attribute over a totally-ordered domain. To handle multiple totally-ordered  attributes $Y_1, \ldots, Y_k$ we will repair each attribute individually. 

We define a ``median'' distribution $A$ 
in terms of its quantile function $F^{-1}_A$:   $F_A^{-1}(u) = \text{median\ }_{x \in X} F_x^{-1}(u)$.
The choice of the term ``median'' is not accidental. 
\begin{lemma}
\label{lemma:remov-disp-impact}
  Let $A$ be a distribution such that $ F_A^{-1}(u) = \text{median\ }_{x \in X} F_x^{-1}(u)$.
Then $A$ is also the distribution minimizing $\sum_{x \in X} d(Y_x, C)$ over all distributions $C$, where $d(\cdot, \cdot)$ is the earthmover distance on $\reals$. 
\end{lemma}
\begin{proof}
For any two distributions $P$ and $Q$ on the line, the earthmover distance (using the underlying Euclidean distance $d(x,y) = |x-y|$ as the metric) can be written as 
\[ d(P, Q) = \int_0^1 |F^{-1}_P(u) - F^{-1}_Q(u)| du \]
In other words, the map $P \mapsto F^{-1}_P$ is an isometric embedding of the earthmover distance into $\ell_1$. 

Consider now a set of points $p_1, \ldots, p_n  \in \ell_1$. Their $1$-median -- the point minimizing $\sum_i \|p_i - c\|_1$ -- is the point whose $j^{\text{th}}$ coordinate is the median of the $j^{\text{th}}$ coordinates of the $p_i$. This is precisely the definition of the distribution $A$ (in terms of $F^{-1}_A$). 
\end{proof}

\begin{figure}[tb]
\begin{center}
\includegraphics[width=0.75\linewidth]{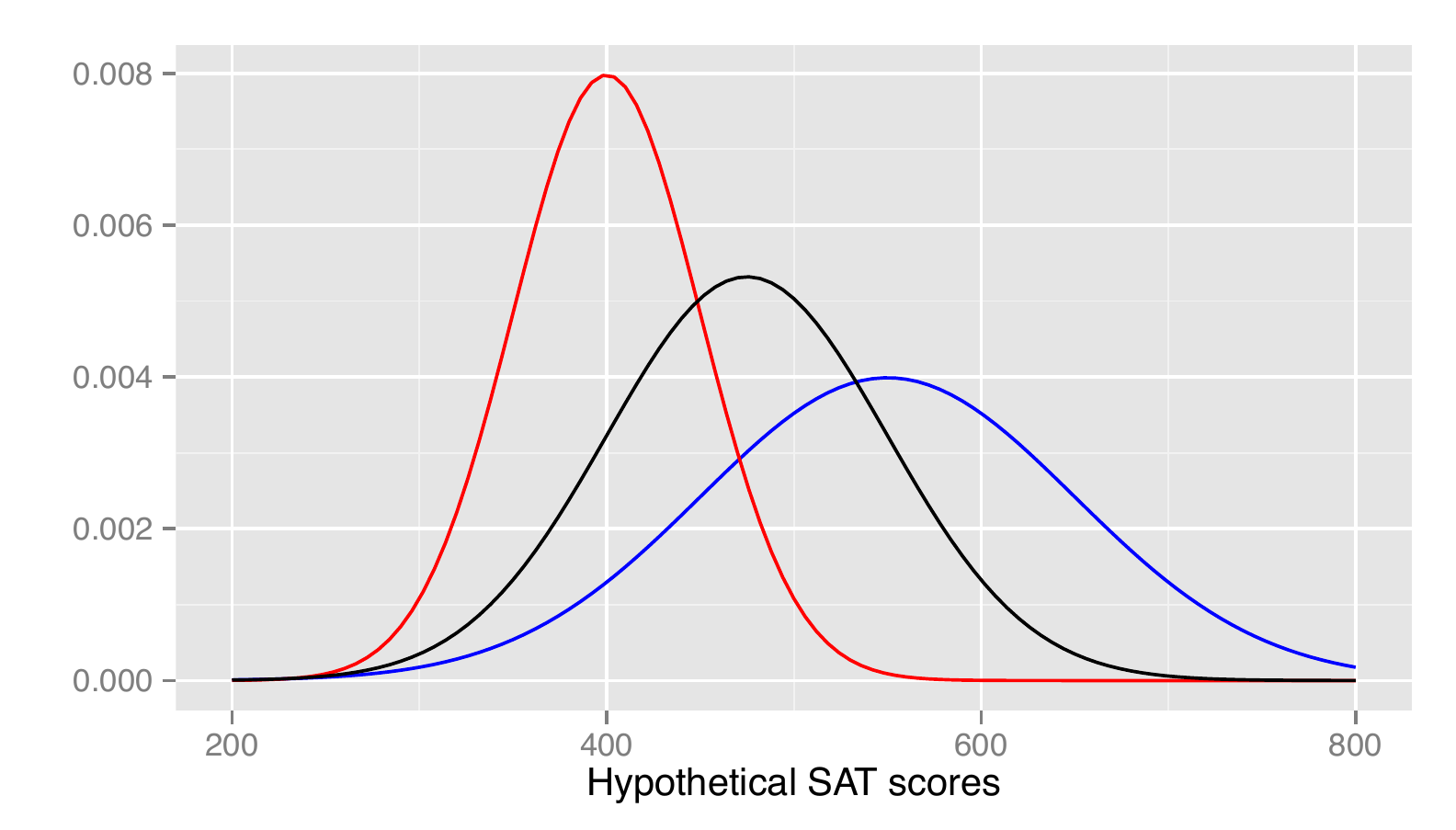}
\caption{Consider the fake probability density functions shown here where the blue curve shows the distribution of SAT scores ($Y$) for $X = \texttt{female}$, with $\mu = 550,\sigma = 100$, while the red curve shows the distribution of SAT scores for $X = \texttt{male}$, with $\mu = 400,\sigma = 50$.  The resulting fully repaired data is the distribution in black, with $\mu = 475,\sigma = 75$.  Male students who originally had scores in the 95th percentile, i.e., had scores of 500, are given scores of 625 in the 95th percentile of the new distribution in $\bar{Y}$, while women with scores of 625 in $\bar{Y}$ originally had scores of 750.}
\label{fig:repair}
\end{center}
\end{figure}

\mypara{Algorithm.} Our repair algorithm creates $\bar{Y}$, such that for all $y \in Y_x$, the corresponding $\bar{y} = F_A^{-1}(F_x(y))$.  The resulting $\bar{D} = (X, \bar{Y}, C)$ changes only $Y$ while the protected attribute and class remain the same as in the original data, thus preserving the ability to predict the class.  See Figure \ref{fig:repair} for an example.

\mypara{Notes.}
\label{sec:notes-1}
We note that this definition is reminiscent of the method by which partial rankings are combined to form a total ranking. The rankings are ``Kemeny''-ized by finding a ranking that minimizes the sum of distances to the original rankings. However, there is a crucial difference in our procedure. Rather than merely reorganizing the data into a total ranking, we are modifying the data to construct this consensus distribution. 

\begin{theorem}
$\bar{D}$ is fair and strongly preserves rank.
\end{theorem}

\begin{proof}
In order to show that $\bar{D}$ strongly preserves rank, recall that we would like to show that $F_x(y) = F_x(\bar{y})$ for all $x \in X$, $\bar{y} \in \bar{Y}_x$, and $y \in Y_x$.  Since, by definition of our algorithm, $\bar{y} = F_A^{-1}(F_x(y))$, we know that
$F_x(\bar{y}) = F_x( F^{-1}_A (F_x(y))) \mbox{,} $
so we would like to show that $F_x(F_A^{-1}(z)) = z$ for all $z \in [0,1]$ and for all $x$.  Recall that $F_A^{-1}(z) = \text{median\,}_{x \in X}F_x^{-1}(z)$. 

Suppose the above claim is not true. Then there are two values $z_1 < z_2$ and some value $x$ such that  $F_x(F_A^{-1}(z_1)) > F_x(F_A^{-1}(z_2))$. That is, there is some $x$ and two elements $y_1 = F_A^{-1}(z_1)$, $y_2 = F_A^{-1}(z_2)$ such that $y_1 > y_2$. Now we know that $y_1 = \text{median\,}_{x \in X} F_x^{-1}(z_1)$. Therefore, if $y_1 > y_2$ it must be that there are strictly less than $|X|/2$ elements of the set $\{ F_x^{-1}(z_1) | x \in X\}$ below $y_2$. But by the assumption that $z_1 < z_2$, we know that  each element of $\{ F_x^{-1}(z_1) | x \in X\}$ is above the corresponding element of $\{ F_x^{-1}(z_2) | x \in X\}$ and there are $|X|/2$ elements of this latter set below $y_2$ by definition. Hence we have a contradiction and so a flip cannot occur, which means that the claim is true. 

Note that the resulting  $\bar{Y}_x$ distributions are the same for all $x \in X$, so there is no way for Bob to differentiate between the protected attributes. Hence the algorithm is $1$-fair. 
\end{proof}

This repair has the effect that if you consider the $\bar{Y}$ values at some rank $z$, the probability of the occurrence of a data item with attribute $x \in X$ is the same as the probability of the occurrence of $x$ in the full population.  This informal observation gives the intuitive backing for the lack of predictability of $X$ from $\bar{Y}$ and, hence, the lack of disparate impact in the repaired version of the data.

\subsection{Partial Repair}
Since the repair process outlined above is likely to degrade Alice's ability to classify accurately, she might want a \emph{partially repaired} data set instead.  This in effect creates a tradeoff between the ability to classify accurately and the fairness of the resulting data. This tradeoff can be achieved by simply moving each inverse quantile distribution only part way towards the median distribution.  Let $\lambda \in [0,1]$ be the amount of repair desired, where $\lambda = 0$ yields the unmodified data set and $\lambda = 1$ is the fully repaired version described above.  Recall that $F_x: Y_x \rightarrow [0,1]$ is the function giving the rank of $y$.  The repair algorithm for $\lambda = 1$ creates $\bar{Y}$ such that $\bar{y} = F_{A}^{-1}(F_x(y))$ where $A$ is the median distribution.  

In the partial repair setting we will be creating a different distribution $A_x$ for each protected value $x \in X$ and setting $\bar{y} = F_{A_x}^{-1}(F_x(y))$. Consider the ordered set of all  $y$ at rank $u$ in their respective conditional distributions i.e the set $U(u) = \{ F_x^{-1}(u) | x \in X \}$. We can associate with $U$ the cumulant function $UF(u, y) = |\{ y' \ge y | y \in U(u) \}| / |U(u)|$ and define the associated quantile function $UF^{-1}(u, \alpha) = y$ where $UF(u, y) = \alpha$. We can restate the full repair algorithm in this formulation as follows: for any $(x,y)$,  $\bar{y} = UF^{-1}(F_x(y), 1/2)$. 

We now describe two different approaches to performing a partial repair, each with their own advantages and disadvantages. Intuitively, these repair methods differ in which space they operate in: the \emph{combinatorial} space of ranks or the \emph{geometric} space of values. 
\subsubsection{A Combinatorial Repair}
\label{sec:combinatorial-repair}

The intuition behind this repair strategy is that each item, rather than being moved to the median of its associated distribution, is only moved part of the way there, with the amount moved being proportional (in rank) to its distance from the median. 
\begin{defn}[Combinatorial Repair]
  Fix an $x$ and consider any pair $(x,y)$. Let $r = F_x(y)$ be the rank of $y$ conditioned on $X = x$. Suppose that in the set $U(r)$ (the collection of all $y' \in Y$ with rank $r$ in their respective conditional distributions) the rank of $y$ is $\rho$. Then we  replace $y$ by $\bar{y} \in U(r)$ whose rank in $U(r)$ is $\rho' = \lfloor (1-\lambda)\rho + \lambda/2\rfloor$. Formally, $\bar{y} = UF^{-1}(r,\rho')$. We call the resulting data set $\bar{D}_\lambda$.  
\end{defn}
While this repair is intuitive and easy to implement, it does not satisfy the property of strong rank preservation. In other words, it is possible that two pairs $(x, y)$ and $(x, y')$ with $y > y'$ to be repaired in a way that $\bar{y} < \bar{y'}$. While this could potentially affect the quality of the resulting data (we discuss this in Section~\ref{sec:fairness_utility}), it does not affect the fairness properties of the repair. Indeed, we formulate the fairness properties of this repair as a formal conjecture. 

\begin{conjecture}
\label{partial_eps_fair}
$\bar{D}_\lambda$ is $f(\lambda)$-fair for a monotone function $f$. 
\end{conjecture}


\subsubsection{A Geometric Repair}
\label{sec:geometric-repair}

The algorithm above has an easy-to-describe \emph{operational} form. It does not however admit a \emph{functional} interpretation as an optimization of a certain distance function, like the full repair. For example, it is not true that for $\lambda = 1/2$ the modified distributions $\bar{Y}$ are equidistant (under the earthmover distance) between the original unrepaired distributions and the full repair.  The algorithm we propose now does have this property, as well as possessing a simple operational form. The intuition is that rather than doing a linear interpolation in \emph{rank space} between the original item and the fully repaired value, it does a linear interpolation in the original \emph{data space}. 

\begin{defn}[Geometric Repair]
Let $F_A$ be the cumulative distribution associated with $A$, the result performing a full repair on the conditional cumulative distributions as described in Section~\ref{sec:repairing-data}. 
Given a conditional distribution $F_x(y)$, its $\lambda$-partial repair is given by 
\[ \bar{F}^{-1}_x(\alpha) = (1-\lambda) F^{-1}_x(\alpha) + \lambda (F_A)^{-1}(\alpha) \]
\end{defn}

Linear interpolation allows us to connect this repair to the underlying earthmover distance between repaired and unrepaired distributions. In particular, 

\begin{theorem}
For any $x$, $ d(Y_x, \bar{Y}_x) = \lambda d(Y_x, Y_A) $
where $Y_A$ is the distribution on $Y$ in the full repair, and $\bar{Y}_x$ is the $\lambda$-partial repair. Moreover, the repair strongly preserves rank.
\end{theorem}
\begin{proof}
The earthmover distance bound follows from the proof of Lemma \ref{lemma:remov-disp-impact} and the isometric mapping between the earthmover distance between $Y_x$ and $\bar{Y}_x$ and the $\ell_1$ distance between $F_x$ and $\bar{F}_x$. Rank preservation follows by observing that the repair is a \emph{linear} interpolation between the original data  and the full repair (which preserves rank by Lemma~\ref{lemma:remov-disp-impact}). 
\end{proof}

\subsection{Fairness / Utility Tradeoff}
The reason partial repair may be desired is that increasing fairness may result in a loss of utility.  Here, we make this intuition precise.  Let $\bar{D}_\lambda = (X, \bar{Y}, C)$ be the partially repaired data set for some value of $\lambda \in [0,1]$ as described above (where $\bar{D}_{\lambda = 0} = D$).  Let $\bar{g}_\lambda \colon \bar{Y} \to C$ be the classifier with the utility we are trying to measure.  
\begin{defn}[Utility]
\label{def:utility}
The \emph{utility} of a classifier $\bar{g}_\lambda \colon \bar{Y} \to C$ with respect to some partially repaired data set $\bar{D}_\lambda$ is
\[ \gamma(\bar{g}_\lambda, \bar{D}_\lambda) = 1 - \ber(\bar{g}_\lambda(\bar{y}), c) .\]
\end{defn}
If the classifier $\bar{g}_{\lambda=0} \colon Y \to C$ has an error of zero on the unrepaired data, then the utility is 1.  More commonly, $\gamma(\bar{g}_{\lambda=0}, \bar{D}_{\lambda=0}) < 1$.  In our experiments, we will investigate how $\gamma$ decreases as $\lambda$ increases.


%% file: experiments.tex
\section{Experiments}
\label{sec:experiments}
We will now consider the certification algorithm and repair algorithm's fairness/utility tradeoff experimentally on three data sets.  The first is the \emph{Ricci data set} at the heart of the Ricci v. DeStefano case \cite{Ricci09}.  It consists of 118 test taker entries, each including information about the firefighter promotion exam taken (Lieutenant or Captain), the score on the oral section of the exam, the written score, the combined score, 
and the race of the test taker (black, white, or Hispanic).  In our examination of the protected race attribute, we will group the black and Hispanic test takers into a single non-white category.  The classifier originally used to determine which test takers to promote was the simple threshold classifier that allowed anyone with a combined score of at least 70\% to be eligible for promotion \cite{Miao11RicciStats}.  Although the true number of people promoted was chosen from the eligible pool according to their ranked ordering and the number of slots available, for simplicity in these experiments we will describe all eligible candidates as having been promoted.  We use a random two-thirds / one-third split for the training / test data.

\looseness-1 The other two data sets we will use are from the UCI Machine Learning Repository\footnote{\url{http://archive.ics.ucu.edu/ml}}.  So that we can compare our results to those of Zemel et al. \cite{icml2013_zemel13}, we will use the same data sets and the same decisions about what constitutes a sensitive attribute as they do.  First, we will look at the \emph{German credit} data set, also considered by  Kamiran and Calders \cite{Kamiran09Classifying}.  It contains 1000 instances, each of which consists of 20 attributes and a categorization of that instance as GOOD or BAD.  The protected attribute is Age.  In the examination of this data set with respect to their discriminatory measure, Kamiran and Calders found that the most discrimination was possible when splitting the instances into YOUNG and OLD at age 25 \cite{Kamiran09Classifying}.  We will discretize the data accordingly to examine this potential worst case.  We use a random two-thirds / one-third split for the training / test data.

We also look at the \emph{Adult income} data set, also considered by Kamishima et al. \cite{Kamishima11Fairness}.  It contains 48,842 instances, each of which consists of 14 attributes and a categorization of that person as making more or less than \$50,000 per year.  The protected attribute we will examine is Gender.  Race is also an attribute in the data, and it will be excluded for classification purposes, except for when examining the effects of having multiple protected attributes - in this case, race will be categorized as white and non-white. The training / test split given in the original data is also used for our experiments.

For each of these data sets, we look at a total of 21 versions of the data - the original data set plus 10 partially or fully repaired attribute sets for each of the combinatorial and geometric partial repairs.  These are the repaired attributes for $\lambda \in [0,1]$ at increments of $0.1$.  Data preprocessing was applied before the partial repair algorithm was run.

\vspace*{0.1in}\mypara{Preprocessing.} Datasets were preprocessed as follows:
\begin{enumerate}
\item Remove all protected attributes from $Y$. This ensures that we are not trying to learn a classifier that depends on other protected attributes that might correlate with the target protected attribute.  (The repair process does still get to know $X$.)
\item Remove all unordered categorical features since our repair procedure assumes that the space of values is ordered.  Ordered categories are converted to integers.\footnote{On the Adult Income data, it happens that all missing values were of these unordered categorical columns, so no data sets had missing values after this step.}
\item Scale each feature so that the minimum is zero and the maximum is one.
\end{enumerate}

\vspace*{0.1in}\mypara{Classifiers.} Three different classifiers were used as oracles for measuring discrimination (under the disparate impact measure and a measure by Zemel et al. \cite{icml2013_zemel13}), and to test the accuracy of a classification after repair.  The classifiers used for our experimental tasks were provided by the Scikit-learn\footnote{\url{http://scikit-learn.org}.} python package. 
\begin{itemize}
\item [LR:] Logistic Regression: Liblinear's~\cite{fan2008liblinear} logistic regression algorithm for L2 regularization and logistic loss. The classifier was configured to weight the examples automatically so that classes were weighted equally.
\item [SVM:] Support Vector Machine: Liblinear's~\cite{fan2008liblinear} linear SVM algorithm for L2 regularization and L2 loss. The classifier was configured to weight the examples automatically so that classes were weighted equally.
\item [GNB:] Gaussian Na\"ive Bayes: Scikit-Learn's na\"ive Bayes algorithm with a balanced class prior. 
\end{itemize}

\vspace*{0.1in}\mypara{Parameter selection and cross-validation.}
LR and SVM classifiers were cross-validated using three-fold cross validation and the best parameter based on BER was chosen. 
We cross-validated the parameter controlling the tradeoff between regularization and loss, and $13$ parameters between $10^{-3}$ and $10^3$, with logarithms uniformly spaced, were searched.

\vspace*{0.1in}\mypara{Repair details.}
The repair procedure requires a ranking of each attribute.  The numeric values and ordered categorical attributes were ordered in the natural way and then quantiles were used as the ranks.  Since the repair assumes that there is a point at each quantile value in each protected class, the quantiles were determined in the following way.  For each attribute, the protected class with the smallest number of members was determined.  This size determined how many quantile buckets to create.  The other protected classes were then appropriately divided into the same number of quantile buckets, with the median value in each bucket chosen as a representative value for that quantile.  Each quantile value in the fully repaired version is the median of the representative values for that quantile.  The combinatorial partial repair determines all valid values for an attribute and moves the original data part way to the fully repaired data within this space.  The geometric repair assumes all numeric values are allowed for the partial repair.

\subsection{Certification}
The goal in this section is to experimentally validate our certification algorithm, described in \autoref{sec:certifying-lack-of}.  On each of the data sets described above, we attempt to predict the protected attribute from the remaining attributes.  The resulting \ber is compared to $\di(g)$ where $g: Y \to C$, i.e., the disparate impact value as measured when some classifier attempts to predict the class given the non-protected attributes.  From the underlying data, we can calculate the \ber threshold $\epsilon = 1/2 - \beta / 8$.  Above this threshold, any classifier applied to the data will have disparate impact.  The threshold is chosen conservatively so as to preclude false positives (times when we falsely declare the data to be safe from disparate impact).

\begin{figure}
\begin{center}
\includegraphics[width=0.75\linewidth]{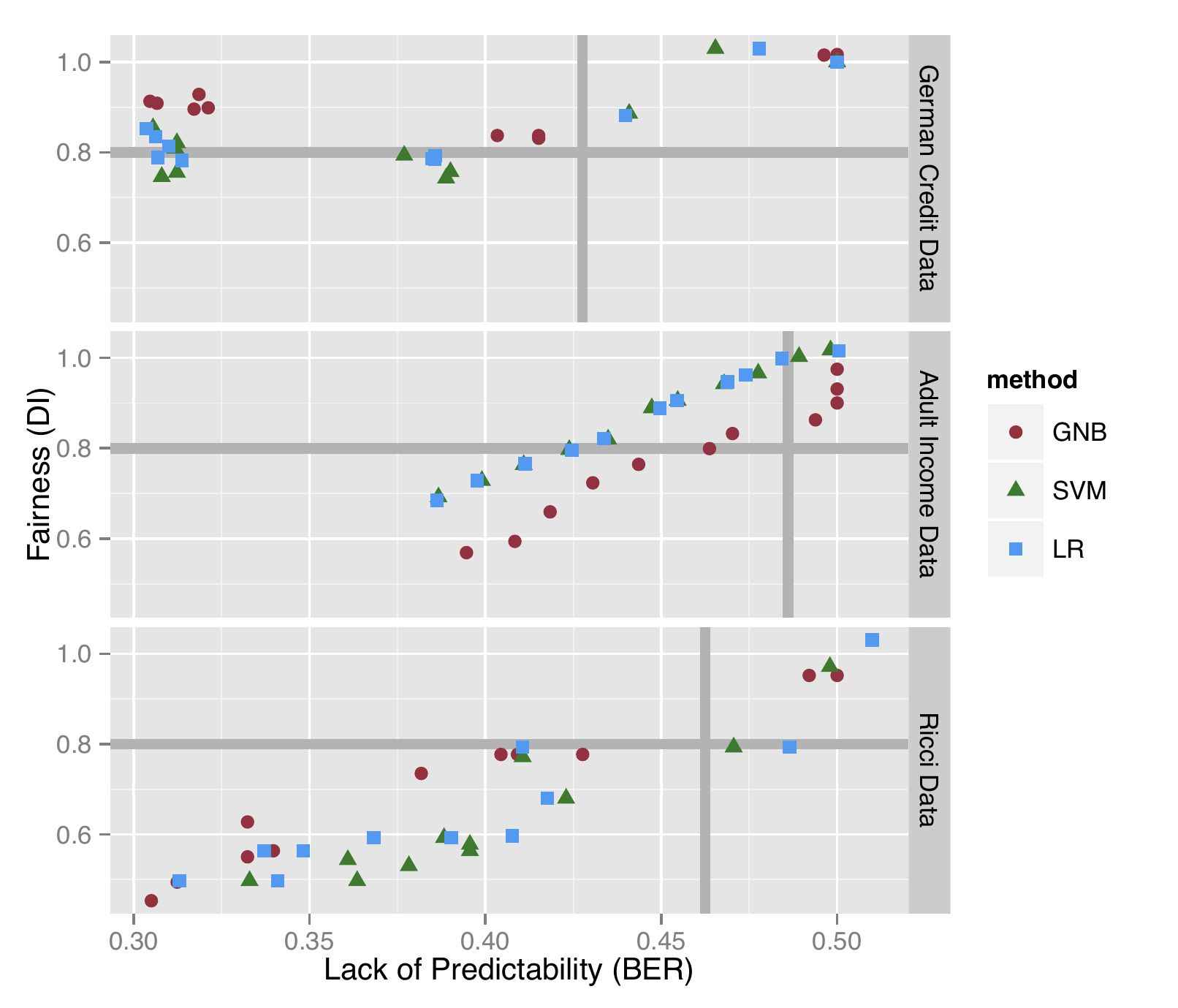}
\caption{Lack of predictability (\ber) of the protected attributes on the German Credit Adult Income, and Ricci data sets as compared to the disparate impact found in the test set when the class is predicted from the non-protected attributes.  The certification algorithm guarantees that points to the right of the \ber threshold are also above $\tau = 0.8$, the threshold for legal disparate impact. For clarity, we only show results using the combinatorial repair, but the geometric repair results follow the same pattern. }
\label{fig:predictability_di}
\end{center}
\end{figure}

\looseness-1 In Figure \ref{fig:predictability_di} we can see that there are no data points greater than the \ber threshold and also much below $\tau = 0.8$, the threshold for legal disparate impact.  The only false positives are a few points very close to the line.  This is likely because the $\beta$ value, as measured from the data, has some error.  We can also see, from the points close to the \ber threshold line on its left but below $\tau$ that while we chose the threshold conservatively, we were not overly conservative.  Still, using a classifier other than the purely biased one in the certification algorithm analysis might allow this threshold to be tightened.

\looseness-1 The points in the upper left quadrant of these charts represent false negatives of our certification algorithm on a specific data set and a specific classifier.  However, our certification algorithm guarantees lack of disparate impact over \emph{any} classifier, so these are not false negatives in the traditional sense.  In fact, when a single data set is considered over all classifiers, we see that all such data sets below the \ber threshold have some classifier that has \di close to or below $\tau = 0.8$.

One seemingly surprising artifact in the charts is the vertical line in the Adult Income data chart at $\ber=0.5$ for the GNB repair.  Recall that the chart is based off of two different confusion matrices - the $\ber$ comes from predicting gender while the disparate impact is calculated when predicting the class.  In a two class system, the $\ber$ cannot be any higher than $0.5$, so while the ability to predict the gender cannot get any worse, the resulting fairness of the class predictions can still improve, thus causing the vertical line in the chart.

\subsection{Fairness / Utility Tradeoff}
\label{sec:fairness_utility}

The goal in this section is to determine how much the partial repair procedure degrades utility.  Using the same data sets as described above, we will examine how the utility (see Definition \ref{def:utility}) changes \di (measuring fairness) increases.  Utility will be defined with respect to the data labels.  Note that this may itself be faulty data, in that the labels may not themselves provide the best possible utility based on the underlying, but perhaps unobservable, desired outcomes.  For example, the results on the test from the Ricci data may not perfectly measure a firefighter's ability and so outcomes based on that test may not correctly predict who should be promoted.  Still, in the absence of knowledge of more precise data, we will use these labels to measure utility.  For the Ricci data, which is unlabeled, we will assume that the true labels are those provided by the simple threshold classifier used on the non-repaired version of the Ricci data, i.e. that anyone with a score of at least 70\% should pass the exam.  Disparate impact (\di) for all data sets is measured with respect to the predicted outcomes on the test set as differentiated by protected attribute.  The SVM described above is used to classify on the Adult Income and German Credit data sets while the Ricci data uses the simple threshold classifier.  The utility ($1-\ber$) shown is based on the confusion matrix of the original labels versus the labels predicted by these classifiers.

\begin{figure}
  \begin{center}
    \includegraphics[width=0.75\linewidth]{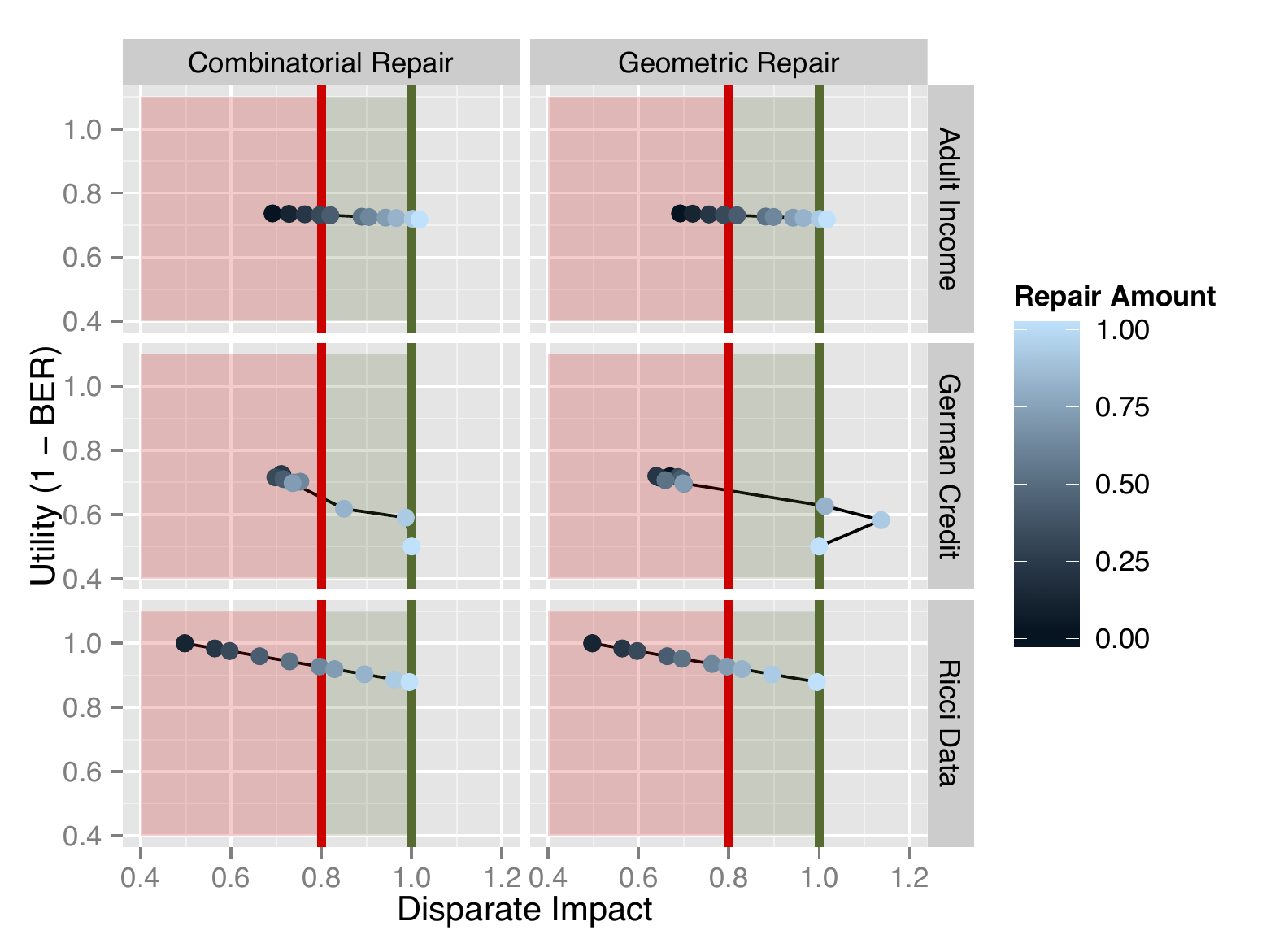}
    \caption{Disparate impact (\di) vs. utility (1-\ber ) from our combinatorial and geometric partial repair processes using the SVM to classify on the Adult Income and German Credit data sets and the simple threshold classifier on the Ricci data set.  Recall that only points with $\di \geq \tau = 0.8$ are legal.  $\di =1.0$ represents full fairness.}
    \label{fig:fairness_utility}
  \end{center}
\end{figure}

The results, shown in Figure \ref{fig:fairness_utility}, demonstrate the expected decay over utility as fairness increases.  Each unrepaired data set begins with $\di < 0.8$, i.e., it would fail the 80\% rule, and we are able to repair it to a legal value.  For the Adult Income data set, repairing the data fully only results in a utility loss from about 74\% to 72\%, while for the German Credit data, repairing the data fully reduces the utility from about 72\% to 50\% - essentially random.  We suspect that this difference in decay is inherent to the class decisions in the data set (and the next section will show that other existing fairness repairs face this same decay).  We suspect that the lack of linearity in the utility decay in the German Credit data after it has fairness greater than $\di = 0.8$ is due to this low utility.

Looking more closely at the charts, we notice that some of the partially repaired data points have $\di > 1$.  Since \di is calculated with respect to fixed majority and minority classes, this happens when the classifier has given a good outcome to proportionally more minority than majority class members.  These points should be considered unfair to the majority class.

Figure \ref{fig:fairness_utility} also shows that combinatorial and geometric repairs have similar \di and utility values for all partial repair data sets.  This means that either repair can be used.

\looseness-1 \vspace*{0.1in}\mypara{Multiple Protected Attributes.}
Our repair procedure can operate over the joint distribution of multiple protected attributes.  To examine how this affects utility, we considered the Adult Income data set repaired by gender only, race only, and over both gender and race.  
For the repairs with respect to race, a binary racial categorization of white and non-white is used.  Repairs with respect to both race and gender are taken over the joint distribution.  In the joint distribution case, the \di calculated is the average of the \di of each of the three protected sets (white women, non-white men, and non-white women) with respect to the advantaged group (white men).  The classifier used to predict the class from the non-protected attributes is the SVM described earlier.

\begin{figure}
\begin{center}
\includegraphics[width=0.75\linewidth]{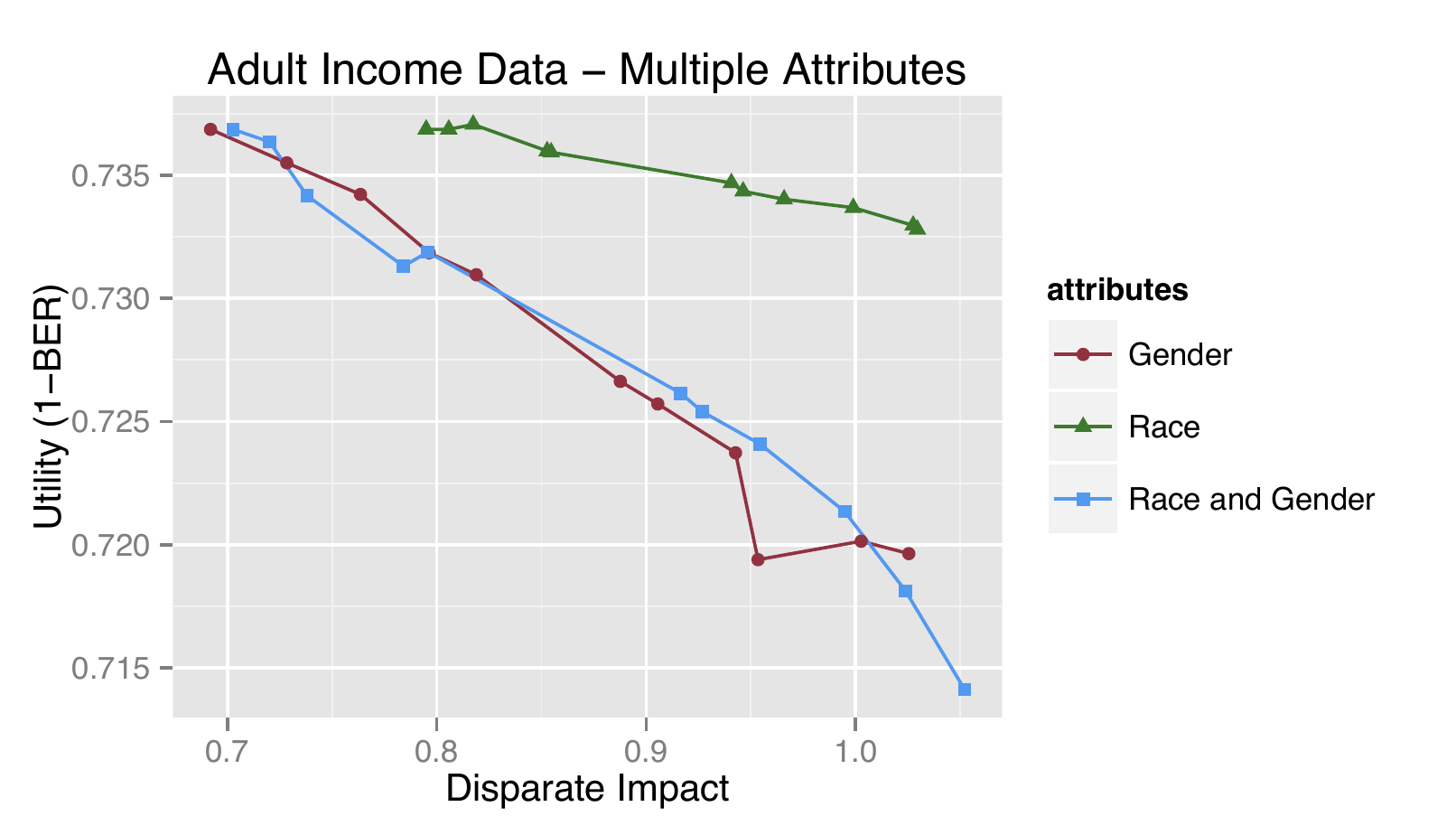}
\caption{Disparate impact (\di) vs. utility (1-\ber ) from our combinatorial and geometric partial repair processes using the SVM as the classifier.  For clarity in the figure, only the combinatorial repairs are shown, though the geometric repairs follow the same pattern.}
\label{fig:fairness_utility_multiple}
\end{center}
\end{figure}

\looseness-1 The results, shown in Figure \ref{fig:fairness_utility_multiple}, show that the utility loss over the joint distribution is close to the maximum of the utility loss over each protected attribute considered on its own.  In other words, the loss does not compound.  These good results are likely due in part to the size of the data set allowing each subgroup to still be large enough.  On such data sets, allowing all protected attributes to be repaired appears reasonable.

\subsection{Comparison to previous work}

\begin{figure}
  \begin{center}
    \includegraphics[width=0.75\linewidth]{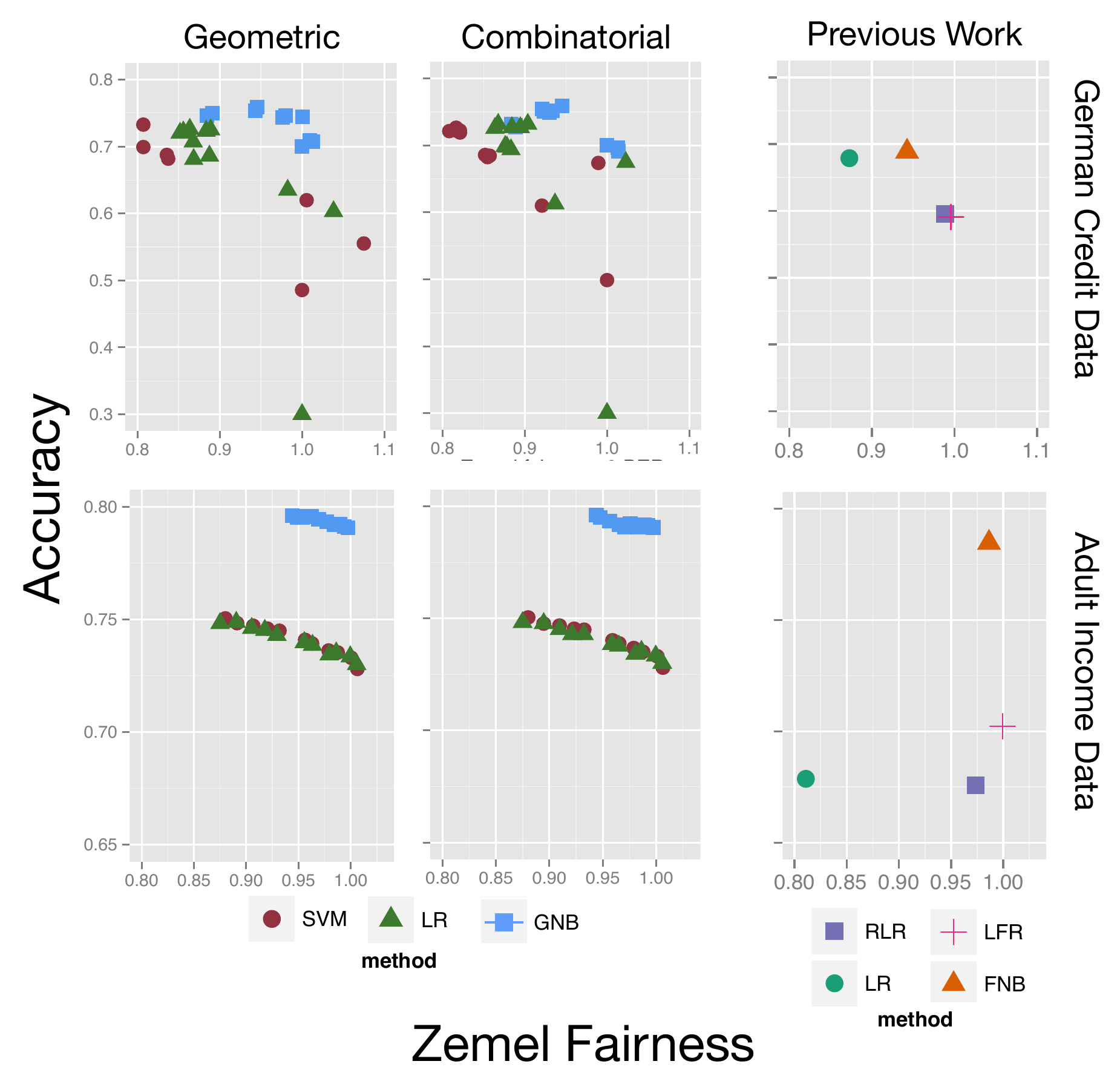}
    \caption{Zemel fairness vs. accuracy from our combinatorial and geometric partial repairs as compared to previous work. Legend: RLR, Regularized Logistic Regression \cite{Kamishima11Fairness}; LFR, Learned Fair Representations \cite{icml2013_zemel13}; FNB, Fair Na\"{i}ve Bayes \cite{Kamiran09Classifying}; GNB, Gaussian Na\"{i}ve Bayes with balanced prior; LR, Logistic Regression; SVM, Support Vector Machine.}
    \label{fig:zemel_accuracy}
  \end{center}
\end{figure}

Here, we compare our results to related work on the German credit data and Adult income data sets.  Logistic regression is used as a baseline comparison, fair naive Bayes is the solution from Kamiran and Calders \cite{Kamiran09Classifying}, regularized logistic regression is the repair method from Kamishima et al. \cite{Kamishima11Fairness}, and learned fair representations is Zemel et al.'s solution \cite{icml2013_zemel13}.  All comparison data is taken from Zemel et al.'s implementations \cite{icml2013_zemel13}.  Zemel et al. define discrimination as $(1-\alpha) - \beta$.  So that increasing Zemel scores mean that fairness has increased, as is the case with \di, we will look at the \emph{Zemel fairness} score which we define as $1 - ((1-\alpha) - \beta) = 2 \cdot \ber$.  Accuracy is the usual rate of successful classification.  Unlike the compared works, we do not choose a single partial repair point.  Figure \ref{fig:zemel_accuracy} shows our fairness and accuracy results for both combinatorial and geometric partial repairs for values of $\lambda \in [0,1]$ at increments of $0.1$ using all three classifiers described above.

Figure \ref{fig:zemel_accuracy} shows that our method can be flexible with respect to the chosen classifier.  Since the repair is done over the data, we can choose a classification algorithm appropriate to the data set.  For example, on the Adult Income data set the repairs based on Na\"{i}ve Bayes have better accuracy at high values of fairness than the repairs based on Logistic Regression.  On the German and Adult data sets our results show that for any fairness value a partially repaired data set at that value can be chosen and a classifier applied to achieve accuracy that is better than competing methods.

Since the charts in Figure \ref{fig:zemel_accuracy} include unrepaired data, we can also separate the effects of our classifier choices from the effects of the repair.  In each classifier repair series, the data point with the lowest Zemel fairness (furthest to the left) is the original data.  Comparing the original data point when the LR classifier was used to the LR classifier used by Zemel et al. as a comparison baseline, we see a large jump in both fairness and accuracy.  Configuring the classifier to weight classes equally may have accounted for this improvement.


%% file: discussion2.tex
\section{Limitations and Future Work}

Our experiments show a substantial difference in the performance of
our repair algorithm depending on the specific algorithms we
chose. Given the myriad classification algorithms used in practice,
there is a clear need for a future systematic study of the
relationship between dataset features, algorithms, and repair
performance.


In addition, our discussion of disparate impact is necessarily tied to
the legal framework as defined in United States law. It would be
valuable in future work to collect the legal frameworks of different
jurisdictions, and investigate whether a single unifying formulation
is possible.


Finally, we note that the algorithm we present operates only on numerical
attributes. Although we are satisfied with its performance, we chose
this setting mostly for its relative theoretical simplicity. A natural
avenue for future work is to investigate generalizations of our repair
procedures for datasets with different attribute types, such as
categorical data, vector-valued attributes, etc.


%% file: appendix.tex
\appendix

\section{A Survey of Discrimination Types}
Previous work has introduced many seemingly dissimilar notions of discrimination.  Here, we give a categorization of these in terms of disparate treatment or disparate impact.  Note that the notion of disparate impact is \emph{outcome focused}, i.e., it is determined based on a discriminatory outcome and, until a question of the legality of the process comes into play in court, is less concerned with \emph{how} the outcome was determined.  In the cases where previous literature has focused on the process, we will try to explain how we believe these processes would be caught under the frameworks of either disparate impact or disparate treatment, hence justifying both our lack of investigation into processes separately as well as the robustness of the existing legal framework.

\subsection{Disparate Treatment}

Disparate treatment is the legal name given to outcomes that are discriminatory due to choices made explicitly based on membership in a protected class.  In the computer science literature, this has previously been referred to as \textbf{blatant explicit discrimination} \cite{Dwork12Fairness}.  When the protected class is used directly in a model, this has been referred to as \textbf{direct discrimination} \cite{Kamishima12Fairness}.

\paragraph{Reverse tokenism}  Dwork et al. describe a situation in which a strong (under the ranking measure) member of the majority class might be purposefully assigned to the negative class in order to refute a claim of discrimination by members of the protected class by using the rejected candidate as an example \cite{Dwork12Fairness}.  Under a disparate treatment or impact theory, this would \emph{not} be an effective way to hide discrimination, since many qualified members of the majority class would have to be rejected to avoid a claim of \textbf{disparate impact}.  If the rejection of qualified members of the majority class was done explicitly based on their majority class status, this would be categorized as \textbf{disparate treatment} \`{a} la the recent Ricci v. DeStefano decision \cite{Ricci09}.

Note that the Ricci decision specifically focuses on an instance in which disparate impact was found and, as a response, the results of a promotion test were not put into place.  This was found to have disparate treatment against the majority class and presents an interesting circularity that any repair to disparate impact needs to address.

\paragraph{Reduced utility}  Dwork et al. also describe a scenario in which outcomes would be more accurate and more beneficial to the protected class if their membership status was taken into account in the model \cite{Dwork12Fairness}.  Here there are two possible decisions.  If the membership status is explicitly used to determine the outcome then this \emph{may} be \textbf{disparate treatment}, however since the law is largely driven by the cases that are brought, if no individual was unjustly treated, then there might be no-one to bring a case, and so it's possible that explicitly using the protected status would not be considered disparate treatment.

If the membership status is \emph{not} used and the protected class and majority class are not represented in close to equal proportion in the positive outcome, then this is \textbf{disparate impact}.  In either case, Dwork et al. might argue that individual fairness has not been maintained, since even if the protected class is proportionally represented, the wrong \emph{individuals} of that class may be systemically receiving the positive outcome.  It is possible to use the disparate impact framework to test for this as well.  Suppose that $X$ is the protected attribute, with classes $x_1$ and $x_2$.  Suppose that $P$ is the attribute which, when combined with $X$, determined who \emph{should} receive a positive outcome, such that the ``correct" outcomes are positive with values $x_1$ and $p_1$ or $x_2$ and $p_2$.  Then by considering the disparate impact over the classes $(x, p)$ in the joint distribution $(X, P)$, the disparate impact to some subset of population $X$ can be detected.  However, $P$ may be a status that is not legally protected, e.g., if $P$ are the grades of a student applying to college it is legal to take those grades into account when determining their acceptance status, so no illegal disparate impact has occurred.

\subsection{Disparate Impact}

Disparate impact (described formally earlier in this paper), is the legal theory that outcomes should not be different based on individuals' protected class membership, even if the process used to determine that outcome does not explicitly base the decision on that membership but rather on proxy attribute(s).  This idea has been a large point of investigation already in computer science and has appeared under many different names with slightly differing definitions, all of which could be measured as disparate impact.  These other names for disparate impact are listed below.

\paragraph{Redlining}  A historic form of discrimination that uses someone's neighborhood as a proxy for their race in order to deny them services \cite{Dwork12Fairness, Kamishima12Fairness, Calders10NaiveBayes}.

\paragraph{Discrimination Based on Redundant Encoding}  Membership in the protected class is not explicitly used, but that information is encoded in other data that is used to make decisions, resulting in discriminatory outcomes \cite{Dwork12Fairness}.

\paragraph{Cutting off business with a segment of the population in which membership in the protected set
is disproportionately high}  The generalized version of redlining \cite{Dwork12Fairness}.

\paragraph{Indirect Discrimination} A model uses attributes that are not independent of membership in the protected class and generates a discriminatory outcome \cite{Kamishima12Fairness}.

\paragraph{Self-fulfilling prophecy}  
Under this scenario, candidates are brought in for interviews at equal rates based on their class status, but are hired at differing rates.  Dwork et al. state that this may be perceived as fair (by looking solely at the rates of interviews granted) and then could be later used to justify a lower rate of hiring among the protected class based on this historical data \cite{Dwork12Fairness}.  We avoid this problem by concentrating throughout on the \emph{outcomes}, i.e., the percent of each class actually hired.  With this perspective, this issue would be identified using the disparate impact standard.

\paragraph{Negative Legacy}  Kamishima et al. discuss the issue of unfair sampling or labeling in the training data based on protected class status \cite{Kamishima12Fairness}.  If such negative legacy causes the outcomes to differ by membership in the protected class, then such bias will be caught under the disparate impact standard.  If negative legacy does not cause any differences to the outcome, then the bias has been overcome and no discrimination has occurred. 

\paragraph{Underestimation}  Kamishima et al. also discuss the impact of a model that has not yet converged due to the data set's small size as a possible source of discrimination \cite{Kamishima12Fairness}.  Similarly to our understanding of negative legacy, we expect to catch this issue by examining the resulting model's outcomes under the disparate impact standard.

\paragraph{Subset targeting}  Dwork et al. point out that differentially treated protected subgroups could be hidden within a larger statistically neutral subgroup \cite{Dwork12Fairness}.  This is a version of Simpson's paradox \cite{Pearl14Simpson}.  We can find such differential treatment by conditioning on the right protected class subset.  In other words, this is the same as one of our suggestions for how to choose $X$ in the repair process - consider multiple classes over their joint distribution as the protected attribute.